\documentclass[a4paper,11pt,DIV=12]{scrartcl}

\usepackage[utf8]{inputenc}
\usepackage{listings}
\usepackage{color}
\usepackage{amsmath}
\usepackage{stmaryrd}
\usepackage{verbatim}
\usepackage{graphicx}
\usepackage{gb4e}
\usepackage{bussproofs}

\title{Pragmatic Side Effects \\ $\;$ \\ workshop redrawing pragmasemantics borders }
\date{}
\author{Jirka Mar\v{s}\'ik \quad Maxime Amblard}

\begin{document}

\maketitle

\section{Introduction}

In the quest to give a formal compositional semantics to natural languages,
semanticists have started turning their attention to phenomena that have
been also considered as parts of pragmatics (e.g., discourse anaphora and
presupposition projection). To account for these phenomena, the very kinds
of meanings assigned to words and phrases are often revisited. To be more
specific, in the prevalent paradigm of modeling natural language
denotations using the simply-typed lambda calculus (higher-order logic)
this means revisiting the types of denotations assigned to individual parts
of speech.

However, the lambda calculus also serves as a fundamental theory of
computation, and in the study of computation, similar type shifts have been
employed to give a meaning to side effects. Side effects in programming
languages correspond to actions that go beyond the lexical scope of an
expression (a thrown exception might propagate throughout a program, a
variable modified at one point might later be read at an another) or even
beyond the scope of the program itself (a program might interact with the
outside world by e.g., printing documents, making sounds, operating robotic
limbs\ldots).

\section{Side Effects and Pragmatics}

We now explore some of the parallels between side effects of programming
languages and the pragmasemantic phenomena of linguistics.

\subsection{Parallel Functions}

We notice that pragmatics seems to do a similar service to natural language
semantics as does the study of side effects to programming language
semantics. Discourse anaphora is an example of an action whose effect
transcends the lexical scope of the expressions involved (the referent and
the referring expression), similar to the way a mutable store bridges the
gap between a variable write and read instruction. Presuppositions, such as
those triggered by definite descriptions, can be seen as propagating
through the structure of the discourse until they are either validated by
some established or hypothesized knowledge or accomodated at the correct
level, much like an exception is propagating throughout a program until it
is caught by some handler. Finally, pragmatics is interested in how a
linguistic system interacts with the world of its users similar to how
programs interact with the world of their users through side effects.

\subsection{Parallel Theories}

When semanticists turn their attention to phenomena whose effects go beyond
the scope of their syntactic domains, they are often forced to generalize
the types of the denotations assigned by their theory to be able to keep a
compositional treatment. In dynamic semantics, the type of a proposition
changes into a function from discourse contexts to propositions and updated
discouse contexts in order to handle anaphora. In programming languages,
the type of a value changes into a function from states of memory to values
and updated states of memory to handle mutable variables.

Computer scientists have developed general notions of a side effect that
allow us to abstract over effects and compose them with relative ease
(monads and monad morphisms \cite{moggi1990abstract}, algebraic effects and
handlers \cite{plotkin2013handling}). A prominent feature of these theories
is that they decompose a complex denotation type (such as the ones seen
above) into a \emph{computation type} with two components: the type of
value being computed and the set of effects this computation has.

This decomposition allows us to put the effects aside and makes it easier
to explore their combinations. Our motivation is to have grammars that
encompass multiple pragmasemantic phenomena and tackle their interactions,
which haven't been studied as much as the individual phenomena themselves.
So far, we have a prototype dealing with dynamics (based on type-theoretic
dynamic logic \cite{de2006towards}), presuppositions\footnote{By
  presuppositions, we mean the kind of conditions that must be true in
  order for an utterance to be judgeable as either true or false (e.g., the
  presupposition of France having a king in the phrase \emph{the king of
    France is bald}). We haven't covered implicatures in our prototype.}
(based on presuppositions as exceptions \cite{lebedeva2012expression}) and
some of their interactions (the presupposition binding problem).

The denotations we assign are computations, which incur some effects until
they yield some value. We assign to sentences computations that yield
simple propositions (i.e., truth values). These computations can incur side
effects that account for their potential to, e.g., interact with the
anaphoric context or trigger presuppositions. We write the grammar not by
positing what meaning should look like, rather we state what it should
do. In this way, we obtain a contrast between the final value, which is all
about truthiness and which falls straight into the domain of semantics, and
the effects yielded by the interpretation, which include pragmatic
phenomena such as discourse anaphora and presuppositions. This distinction
could thus be seen as a formal incarnation of Stalnaker's distinction
between content and context \cite{stalnaker1974pragmatic}.

\bibliography{effects-paper}
\bibliographystyle{plain}

\end{document}